\documentclass[conference]{IEEEtran}
\usepackage[utf8]{inputenc}
\usepackage[T1]{fontenc}
\usepackage[english]{babel}

\usepackage{amsmath,amssymb,amsfonts}
\usepackage{mathtools}
\usepackage{dsfont}
\usepackage{caption}
\usepackage{amsthm}
\usepackage{makecell}
\usepackage{algorithm}
\usepackage[noend]{algpseudocode} % algorithmicx

\usepackage{graphicx}
\usepackage{float}     % for [H] if truly needed
\usepackage{arydshln}  % ok if you need dashed lines
\usepackage[caption=false,font=footnotesize]{subfig}

\usepackage{xcolor}
\usepackage[hidelinks]{hyperref}

\usepackage[section]{placeins} % for \FloatBarrier

\DeclarePairedDelimiterX{\expectarg}[1]{[}{]}{#1}

\algnewcommand{\LongState}[1]{%
  \State \parbox[t]{\dimexpr\linewidth-\algorithmicindent}{\raggedright #1\strut}%
}
\algnewcommand{\Inputs}[1]{%
  \State \textbf{Inputs:}
  \Statex \hspace*{\algorithmicindent}\parbox[t]{.8\linewidth}{\raggedright #1}}
\algnewcommand{\Initialize}[1]{%
  \State \textbf{Initialize:}
  \Statex \hspace*{\algorithmicindent}\parbox[t]{.8\linewidth}{\raggedright #1}}
\algnewcommand{\Data}[1]{%
  \State \textbf{Data:}
  \Statex \hspace*{\algorithmicindent}\parbox[t]{.8\linewidth}{\raggedright #1}}

\DeclareMathOperator*{\argmin}{\arg\!\min}

\graphicspath{{images/}}

\title{Tricks and Plug-ins for Gradient Boosting in Image Classification}

\author{
Biyi Fang\textsuperscript{1} \quad
Truong Vo\textsuperscript{1} \quad
Jean Utke\textsuperscript{2} \quad
Diego Klabjan\textsuperscript{1} \\
\textsuperscript{1}Northwestern University \quad
\textsuperscript{2}Allstate \\
{\tt\small biyifang2021@u.northwestern.edu} \quad
{\tt\small truongvo2025@u.northwestern.edu} \\
{\tt\small jutke@allstate.com} \quad
{\tt\small d-klabjan@northwestern.edu}
}

\begin{document}
\maketitle

\begin{abstract}
Convolutional Neural Networks (CNNs) have achieved remarkable success across a wide range of machine learning tasks by leveraging hierarchical feature learning through deep architectures. However, the large number of layers and millions of parameters often make CNNs computationally expensive to train, requiring extensive time and manual tuning to discover optimal architectures. In this paper, we introduce a novel framework for boosting CNN performance that integrates dynamic feature selection with the principles of BoostCNN. Our approach incorporates two key strategies-subgrid selection and importance sampling-to guide training toward informative regions of the feature space. We further develop a family of algorithms that embed boosting weights directly into the network training process using a least squares loss formulation. This integration not only alleviates the burden of manual architecture design but also enhances accuracy and efficiency. Experimental results across several fine-grained classification benchmarks demonstrate that our boosted CNN variants consistently outperform conventional CNNs in both predictive performance and training speed.
\end{abstract}

\section{Introduction}
\label{sec:intro}

Deep convolutional neural networks (CNNs) have achieved remarkable success in image representation learning for a wide range of computer vision tasks, including image classification \cite{He2016DeepRL, Krizhevsky2017ImageNetCW, Lin2015BilinearCM}, object detection \cite{Girshick2014RichFH, Iandola2014DenseNetIE, Ren2015FasterRT}, and image segmentation \cite{Girshick2014RichFH, Iandola2014DenseNetIE, Ren2015FasterRT}. However, each vision task typically requires a task-specific architecture, and designing or tuning optimal deep networks remains a computationally intensive challenge. While neural architecture search methods such as AMC \cite{He2018AMCAF} and LEAF \cite{Liang2019EvolutionaryNA} offer automated solutions, they often demand extensive computational resources-ranging from thousands of GPU hours to several weeks of training.

In parallel, ensemble methods such as boosting have gained significant attention for their theoretical and empirical advantages over single-model approaches, especially in structured tasks like decision trees \cite{Quinlan2004InductionOD}. To extend these benefits to CNNs, BoostCNN \cite{Moghimi2016BoostedCN} proposed combining shallow CNNs within a boosting framework, thereby simplifying the architecture design problem. However, this method suffers from high memory usage and long runtimes when the weak learners become moderately complex.

To overcome these limitations, we propose Subgrid BoostCNN, a novel family of boosting algorithms that builds upon BoostCNN by integrating ideas from feature subsetting in random forests. Each weak learner is trained on a dynamically selected subset of image pixels (i.e., a subgrid), reducing computational burden while maintaining representational fidelity. Our subgrid selection is guided by the image gradient and the residual from the boosting iteration, allowing the model to focus on the most informative regions of the image. This approach reduces training complexity by breaking the full spatial pixel dependencies, albeit with the potential trade-off of increased noise.

Additionally, we propose a strategy to further reduce computational cost by avoiding full CNN re-optimization in each boosting iteration. Specifically, we reuse the convolutional layers from the previous learner and pair them with the fixed fully connected classifier from the initial iteration to compute pixel importance. This reused architecture enables us to train new learners efficiently with fewer parameters-requiring only one forward-backward pass per iteration, similar to BoostCNN but with significantly reduced overhead.

In summary, our contributions are as follows:
\begin{itemize}
\item We introduce Subgrid BoostCNN, a boosting-based CNN framework that trains weak learners on dynamically selected subgrids. 
% , substantially reducing training time and memory usage.
\item We develop an efficient architectural reuse mechanism to avoid repeated optimization of full CNNs by leveraging previously trained convolutional backbones and a fixed classifier head.
\item We validate our approach on CIFAR-10, SVHN, and ImageNetSub, showing that Subgrid BoostCNN achieves higher accuracy and lower training time.
% compared to both standard CNNs and BoostCNN, with improvements of up to 12.10\% in accuracy on large-scale datasets.
\end{itemize}

\section{Related Work}

Existing extensions of Gradient Boosting Machines (GBMs) are extensive, so we focus only on those most relevant to our proposed algorithms, along with the related ideas of subgrid and importance sampling. Several works have explored combining boosting with deep CNNs, motivated by the strong representational power of modern convolutional architectures. Early approaches, such as Boosted CNNs with fixed boosted blocks \cite{Brahimi2019BoostedCN} and boosted sampling strategies \cite{Berger2018BoostedTO}, either lack architectural flexibility or treat CNNs as black-box predictors. Other methods use CNN features with AdaBoost \cite{Lee2018ImageCB} or integrate boosting directly into CNNs through incremental boosting layers \cite{Han2016IncrementalBC} or least-squares-based boosting of multiple CNNs \cite{Moghimi2016BoostedCN}, though these often incur high computational cost. More recent hybrid models replace fully connected layers with GBMs \cite{Bui2021GBMCNN} or iteratively add boosted dense layers while freezing earlier weights \cite{gbcnn}. Notably, these methods still train weak learners on all features, whereas importance sampling-well studied in SGD and deep learning contexts for improving convergence \cite{Needell2014StochasticGD, Zhao2015StochasticOW, Katharopoulos2018NotAS, Csiba2018ImportanceSF}-has not been generalized to boosting, leaving an opportunity our work aims to address.

\section{Algorithms}
In this section, we first provide a summary of BoostCNN  \cite{Moghimi2016BoostedCN} and then propose the new algorithm, \textbf{subgrid BoostCNN}, which combines BoostCNN and the subgrid trick.

\subsection{\textbf{Background (BoostCNN):} }
We start with a brief overview of multiclass boosting. Given a sample $x_i\in\mathcal{X}$ and its class label $z_i\in\left\{1,2,\cdots,M\right\}$, multiclass boosting is a method that combines several multiclass predictors $g_t:\mathcal{X}\rightarrow \mathbb{R}^d$ to form a strong committee $f(x)$ of classifiers, i.e. $f(x)=\sum_{t=1}^N\alpha_t g_t(x)$ where $g_t$ and $\alpha_t$ are the weak learner and coefficient selected at the $t^{\mathrm{th}}$ boosting iteration. There are various approaches for multiclass boosting such as \cite{Hastie2009MulticlassA}, \cite{Mukherjee2013ATO}, \cite{Saberian2011MulticlassBT}; we use the GD-MCBoost method of \cite{Saberian2011MulticlassBT}, \cite{Moghimi2016BoostedCN} herein. For simplicity, in the rest of the paper, we assume that $d=M$.

Standard BoostCNN \cite{Moghimi2016BoostedCN} trains a boosted predictor $f(x)$ by minimizing the risk of classification
\begin{align}
\label{risk function}
    \mathcal{R}[f]=\mathrm{E}_{X,Z}\left[L(z,f(x)) \right]\approx \frac{1}{\left|\mathcal{D} \right|}\sum_{(x_i,z_i)\in\mathcal{D}}L(z_i,f(x_i)),
\end{align}
where $\mathcal{D}$ is the set of training samples and 
\begin{align*}
%\label{loss function}
    L(z,f(x))=\sum_{j=1, j\neq z}^M e^{-\frac{1}{2}\left[ \left\langle y_{z},f(x)\right\rangle -\left\langle y_j,f(x) \right\rangle \right]},
\end{align*}
given $y_k=\mathds{1}_k\in\mathbb{R}^M$, i.e. the $k^{\mathrm{th}}$ unit vector. The minimization is via gradient descent in a functional space. Standard BoostCNN starts with $f(x)=\mathbf{0}\in\mathbb{R}^d$ for every $x$ and iteratively computes the directional derivative of risk (\ref{risk function}), for updating $f(x)$ along the direction of $g(x)$
\begin{align}
    \delta \mathcal{R}[f;g]&=\left.\frac{\partial \mathcal{R}[f+\epsilon g]}{\partial \epsilon}\right|_{\epsilon=0}\nonumber\\
    &=-\frac{1}{2\left| \mathcal{D}\right|}\sum_{(x_i,z_i)\in\mathcal{D}}\sum_{j=1}^Mg_j(x_i)w_j(x_i,z_i)\nonumber\\
    &=-\frac{1}{2\left| \mathcal{D}\right|}\sum_{(x_i,z_i)\in\mathcal{D}} g(x_i)^Tw(x_i,z_i),
    \label{functional gradient}
\end{align}
where 
\begin{align}
\label{weight update}
    w_k(x,z)=\left\{\begin{matrix}
&-e^{-\frac{1}{2}\left [ f_{z}(x)-f_k(x) \right ]},\quad k\neq z\\ 
&\sum_{j=1,j\neq k}^M e^{-\frac{1}{2}\left [ f_{z}(x)-f_j(x) \right ]}, \quad k=z,
\end{matrix}\right.
\end{align}
and $g_j(x_i)$ computes the directional derivative along $\mathds{1}_j$. Then, standard BoostCNN selects a weak learner $g^*$ that minimizes (\ref{functional gradient}), which essentially measures the similarity between the boosting weights $w(x_i,z_i)$ and the function values $g(x_i)$. Therefore, the optimal network output $g^*(x_i)$ has to be proportional to the boosting weights, i.e. 
\begin{align}
\label{eq:grad-weight}
    g^*(x_i)=\beta w(x_i, z_i),
\end{align}
for some constant $\beta > 0$. Note that the exact value of $\beta$ is irrelevant since $g^*(x_i)$ is scaled when computing $\alpha^*$. Consequently, without loss of generality, we assume $\beta=1$ and convert the problem to finding a network $g(x)\in \mathbb{R}^M$ that minimizes the square error loss
\begin{align}
\label{weak learner train}
    \mathcal{L}(w,g)=\sum_{(x_i,z_i)\in\mathcal{D}}\left\|g(x_i)-w(x_i,z_i)\right\|^2.
\end{align}
After the weak learner is trained, BoostCNN applies a line search to compute the optimal step size along $g^*$,

\begin{align}
    \alpha^*=\argmin_{\alpha\in\mathbb{R}}\mathcal{R}[f+\alpha g^*].
    \label{boost parameter}
\end{align}

Finally, the boosted predictor $f(x)$ is updated as $f=f+\alpha^* g^*$.
%    \label{boost update}
\vspace{0.3cm}
\subsection{\textbf{Subgrid BoostCNN}}
When considering full-size images, BoostCNN using complex CNNs as weak learners is time-consuming and memory hungry. Consequently, we would like to reduce the size of the images to lower the running time and the memory requirement. A straightforward idea would be downsizing the images directly. A problem of this approach is that the noise would possibly spread out to later learners since a strong signal could be weakened during the downsize process. Another candidate for solving the aforementioned problem is randomly selecting pixels from the original images, however, the fluctuation of the performance of the algorithm would be significant especially when the images are sharp or have a lot of noise. In this paper, we apply the subgrid trick to each weak learner in BoostCNN. The remaining question is how to select a subgrid for each weak learner. Formally, a subgrid is defined by deleting a subset of rows and columns. Moreover, the processed images may not have the same size between iterations, which in turn requires that the new BoostCNN should allow each weak learner to have a at least different dimensions. However, that impedes reusing weak learner model parameters from one  weak learner iterate to next.

In order to address these issues, we first separate a standard deep CNN into two parts. We call all  layers such as convolutional layers and pooling layers, except the last fully-connected (FC) layers, the {\it feature extractor}. In contrast, we call the last FC layers the {\it classifier}. Furthermore, we refer to $g_0$ as the basic weak learner and all the succeeding $g_t$ as the additive weak learners. Subgrid BoostCNN defines an importance index for each pixel $(j,k)$ in the image as 
\begin{align}
\label{importance}
    I_{j,k}=\frac{1}{\left|\mathcal{D}\right|}\sum_{(x_i,z_i)\in\mathcal{D}}\sum_{c\in C}\left|\frac{\partial \mathcal{L}(w,g)}{\partial x_i^{j,k,c}}\right|,
\end{align}
where $x_i^{j,k,c}$ denotes pixel $(j,k)$ in channel $c$ from sample $i$ and $C$ represents the set of all channels. The importance index of a row, column is a summation of the importance indexes in the row, column divided by the number of columns, rows, respectively. This importance index is computed based on the residual of the current predictor. Therefore, a larger importance value means a larger adjustment is needed for this pixel at the current iterate. The algorithm uses the importance index generated based on the feature extractor of the incumbent weak learner and the classifier from $g_0$ to conduct subgrid selection. The selection strategy we apply in the algorithm is deleting less important columns and rows, which eventually provides the important subgrid. After the subgrid is selected, subgrid BoostCNN creates a new tensor $x_i^t$ at iterate $t$, and then feeds it into an appropriate feature extractor followed by a proper classifier. The modified minimization problem becomes
\begin{align}
\label{subgrid weak learner train}
  \mathcal{L}(w,g)=\sum_{(x_i,z_i)\in\mathcal{D}}\left\|g(x_i^t)-w(x_i,z_i)\right\|^2, 
\end{align}
where the modified boosting classifier is 
\begin{align}
    f(x)=\sum_{t=1}^N\alpha_tg_t(x^t).
    \label{subgrid classifier}
\end{align}
In this way, subgrid BoostCNN dynamically selects important subgrids based on the updated residuals. Moreover, subgrid BoostCNN is able to deal with inputs of different sizes by applying different classifiers. Furthermore, we are allowed to pass the feature extractor's parameters from the previous weak learner since the feature extractor is not restricted to the input size. The proposed algorithm (subgrid BoostCNN) is summarized in Algorithm \ref{alg:subBoostCNN}.

\begin{algorithm}[htpb]
  \caption{subgrid BoostCNN}
  \label{alg:subBoostCNN}
  \begin{algorithmic}[1]
  \Inputs{number of classes $M$, number of boosting iterations $N_b$, shrinkage parameter $\nu$, dataset $\mathcal{D}=\left\{(x_1,z_1),\cdots,(x_n,z_n)\right\}$ where $z_i\in\left\{1,\cdots,M\right\}$ is the label of sample $x_i$, and $0<\sigma<1$}
    \Initialize{set $f(x)=\mathbf{0}\in\mathbb{R}^M$, $P_0=\left\{(j,k)\vert (j,k)\mathrm{\,is\,a\,pixel\,in\,}x_i \right\}$ }
    \State compute $w(x_i,z_i)$ for all $(x_i,z_i)$, using (\ref{weight update}) \label{basic start}
    \State train a deep CNN $g_0^*$ to optimize (\ref{weak learner train})\label{basic end}
     \State $f(x) = g_0^*$
    \For{t = $1,2,\cdots$, $N_b$ }
    \State update importance index $I_{j,k}$ for $(j,k)\in P_{t-1}$, using (\ref{importance})\label{generate matrix}
        \State select the subgrid based on $\sigma$ fraction of rows and columns with highest importance index and let $P_t$ be the set of selected pixels; form a new tensor $x_i^t$ for each sample $i$\label{select subgrid}
        \State construct a new proper weak learner architecture\label{construct wl}
        \State compute $w(x_i,z_i)$ for all $i$, using (\ref{weight update}) and (\ref{subgrid classifier})\label{subgrid start}
    \State train a deep CNN $g_t^*$ to optimize (\ref{subgrid weak learner train})
    \State find the optimal coefficient $\alpha_t$, using (\ref{boost parameter}) and (\ref{subgrid classifier})\label{subgrid end}
     \State $f(x) = f(x) + \nu\alpha_t g_t^*$\label{subgrid update}
      \EndFor
      \State \textbf{end for}
  \end{algorithmic}
\end{algorithm}
Subgrid BoostCNN starts by initializing $f(x)=\mathbf{0}\in\mathbb{R}^M$. The algorithm first generates a full-size deep CNN as the basic weak learner, which uses the full image in steps~\ref{basic start}-\ref{basic end}. After the basic weak learner $g_0^*$ is generated, in each iteration, subgrid BoostCNN first updates the importance index $I_{j,k}$ for each pixel $(j,k)$, which has been used in the preceding iterate at step~\ref{generate matrix}. In order to mimic the loss of the full-size image, although we only update the importance indexes for the pixels which have been used in the last iterate, we feed the full-size tensor to the deep CNN $g$ to compute the importance index. The deep CNN $g$ used in (\ref{importance}) to compute the importance value is constructed by copying the feature extractor from the preceding weak learner followed by the classifier in the basic weak learner $g_0^*$. Next, by deleting less important rows and columns based on $I_{j,k}$, which contain $1-\sigma$ fraction of pixels, it finds the most important subgrid having $\sigma$ fraction of pixels at position $P_t$ based on the importance index $I_{j,k}$, and forms a new tensor $x_i^t$ in step~\ref{select subgrid}. Note that $P_t$ is not necessary to be a subset of $P_{t-1}$ and actually is rarely to be a subset of $P_{t-1}$. This only happens when the highest importance index at iterate $t$ is also the highest score at iterate $t-1$. Next, a new additive weak learner is initialized by borrowing the feature extractor from the preceding weak learner $g^*_{t-1}$ followed by a randomly initialized FC layer with the proper size in step~\ref{construct wl}. Once the additive weak learner is initialized, subgrid BoostCNN computes the boosting weights, $w(x)\in\mathrm{R}^M$ according to (\ref{weight update}) and (\ref{subgrid classifier}), trains a network $g_t^*$ to minimize the squared error between the network output and boosting weights using (\ref{subgrid weak learner train}), and finds the boosting coefficient $\alpha_t$ by minimizing the boosting loss (\ref{boost parameter}) in steps~\ref{subgrid start}-\ref{subgrid end}. Lastly, the algorithm adds the network to the ensemble according to $f(x) = f(x) + \nu\alpha_t g_t^*$ for $\nu\in[0,1]$ in step~\ref{subgrid update}.

\section{Experimental Study}
% In this section, we first compare subgrid BoostCNN with standard BoostCNN \cite{Moghimi2016BoostedCN} and deep CNNs. From all of the datasets, we study the performance of the boosting technique, the subgrid trick and the importance sampling strategy. All the algorithms are implemented in Python with PyTorch \cite{paszke2017automatic}. Training is conducted on an NVIDIA Titan XP GPU.

% In this subsection, we illustrate properties of the proposed subgrid BoostCNN and compare its performance with other methods on several image classification tasks. In subgrid BoostCNN, the risk function (\ref{risk function}) we employ is cross entropy, and the input of each weak learner is an image with 3 channels which can be handled by standard Conv2d functions in PyTorch. Meanwhile, we implement the subgrid strategy based on (\ref{importance}) with respect to each pixel $(j,k)$. We delete approximately $10\%$ of the rows and columns, which implies $\sigma = 81\%$ on the total number of pixels, and fix the shrinkage parameter $\nu$ to be $0.02$. In each weak learner, we apply the ADAM algorithm with the learning rate of $0.0001$ and weight decay being $0.0001$.

In this section, we apply the subgrid trick on both standard\textbf{ BoostCNN}~\cite{Moghimi2016BoostedCN} and an \textbf{ensemble of CNNs (e-CNN)}, where multiple CNNs are trained independently and their predictions are averaged without boosting weights or feature selection. We analyze the impact of \textit{boosting}, \textit{subgrid selection}, and \textit{importance sampling}. All models are implemented in PyTorch~\cite{paszke2017automatic} and trained on an NVIDIA Titan XP GPU. Subgrid BoostCNN uses cross-entropy loss and standard 3-channel image inputs handled by \texttt{Conv2d}. The subgrid strategy, based on Equation~(\ref{importance}), drops approximately 10\% of rows and columns, retaining about 81\% of the pixels. We fix the shrinkage parameter to $\nu = 0.02$ and train each weak learner using the ADAM optimizer with a learning rate and weight decay of 0.0001.

We consider CIFAR-10 \cite{Krizhevsky2009LearningML}, SVHN \cite{Netzer2011ReadingDI} and ImageNetSub \cite{Deng2009ImageNetAL} datasets as shown in Table \ref{tb:data1}. For the last dataset, since the original ImageNet dataset is large and takes significant amount of time to train, we select a subset of samples from the original ImageNet dataset. More precisely, we randomly pick 100 labels and select the corresponding samples from ImageNet, which consists of $124,000$ images for training and $10,000$ images for testing. We denote it as ImageNetSub. 
% Data preprocessing consists of three steps: 1. random resizing and cropping with output size $224\times 224$, scale uniformly sampled from [0.08, 1.0] and make the aspect ratio uniformly sampled from [0.75, 1.33]; 2. random horizontal flipping with  flipping probability $0.5$; 3. normalization for each channel.

\begin{table}[H]
\centering
\caption{Image datasets used in our experiments.}
\label{tb:data1}
\resizebox{0.9\columnwidth}{!}{
\begin{tabular}{|l|c|c|}
\hline
Dataset & \makecell{Number of\\Training / Testing} & \makecell{Number of\\Classes} \\ \hline
CIFAR-10    & 50k / 10k   & 10  \\ \hline
SVHN        & 73k / 26k   & 10  \\ \hline
ImageNetSub & 124k / 10k  & 100 \\ \hline
\end{tabular}
}
\end{table}

For training, we employ three different deep CNNs, which are ResNet-18, ResNet-50 and ResNet-101. For each combination of dataset/CNN, we first train the deep CNN for a certain number of epochs, and then initialize the weights in the basic weak learner for the boosting algorithms as the weights in the deep CNN. In the subgrid BoostCNN experiments, we use $10$ CNN weak learners. We train each weak learner for 15 epochs. For comparison, we train the ensemble method where multiple CNNs are trained separately, and their outputs are aggregated to produce a final prediction without boosting weight update and always using all features), denoted by \textbf{e-CNN} and the subgrid ensemble method named as subgrid e-CNN (without boosting weight update in step~\ref{subgrid start} in Algorithm \ref{alg:subBoostCNN}) for 10 iterates as well. Notice that subgrid e-CNN essentially mimics random forests. We also train the single deep CNN for 150 epochs to represent approximately the same computational effort as training 10 CNN weak learners for 15 epochs.

\begin{figure*}[htbp]
\centering
\begin{minipage}{.45\textwidth}
  \centering
  \includegraphics[width=\linewidth]{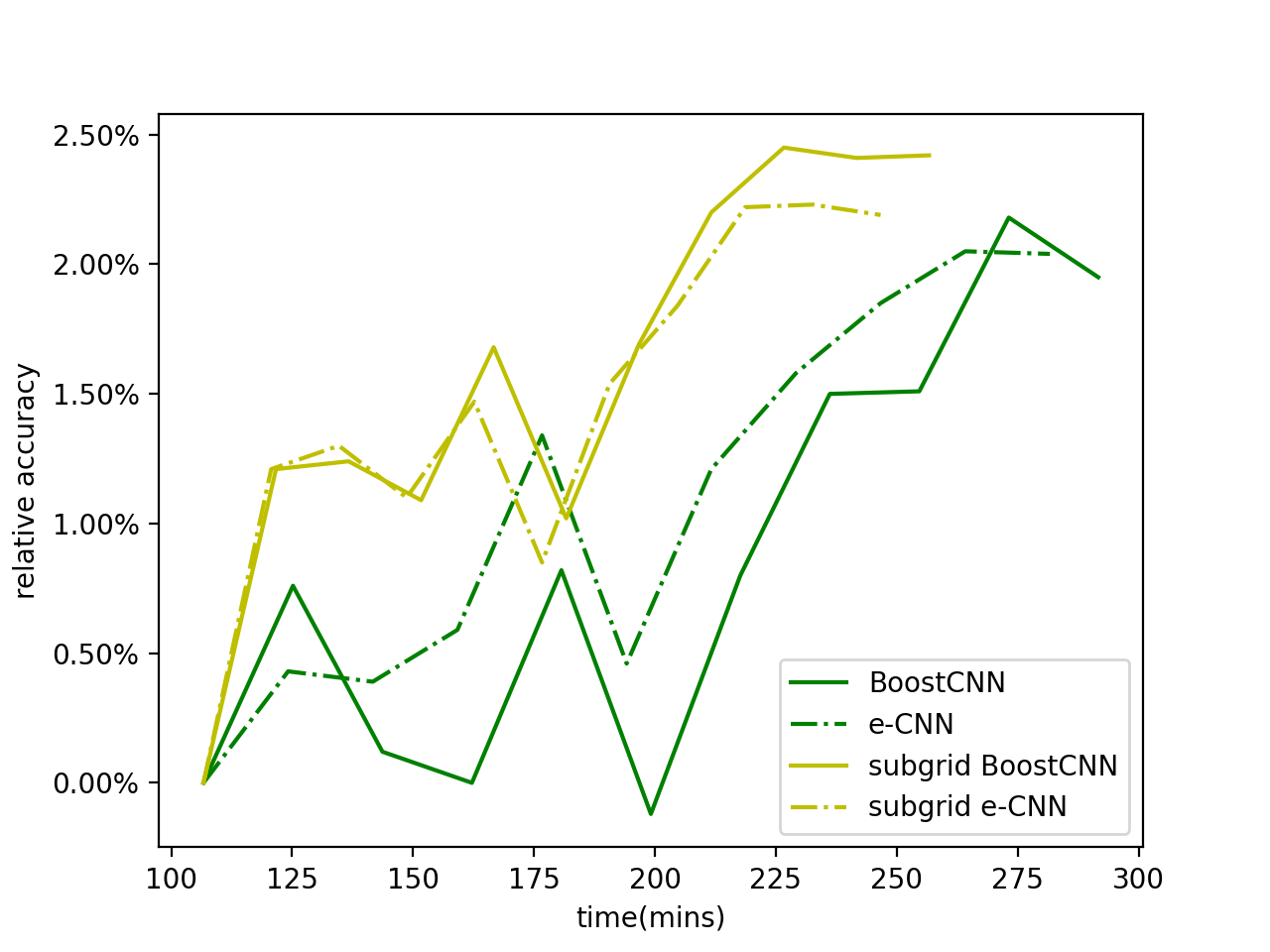}
  \captionof{figure}{ResNet-18 on CIFAR-10}
  \label{fig:18-cifar}
\end{minipage}%
\begin{minipage}{.45\textwidth}
  \centering
  \includegraphics[width=\linewidth]{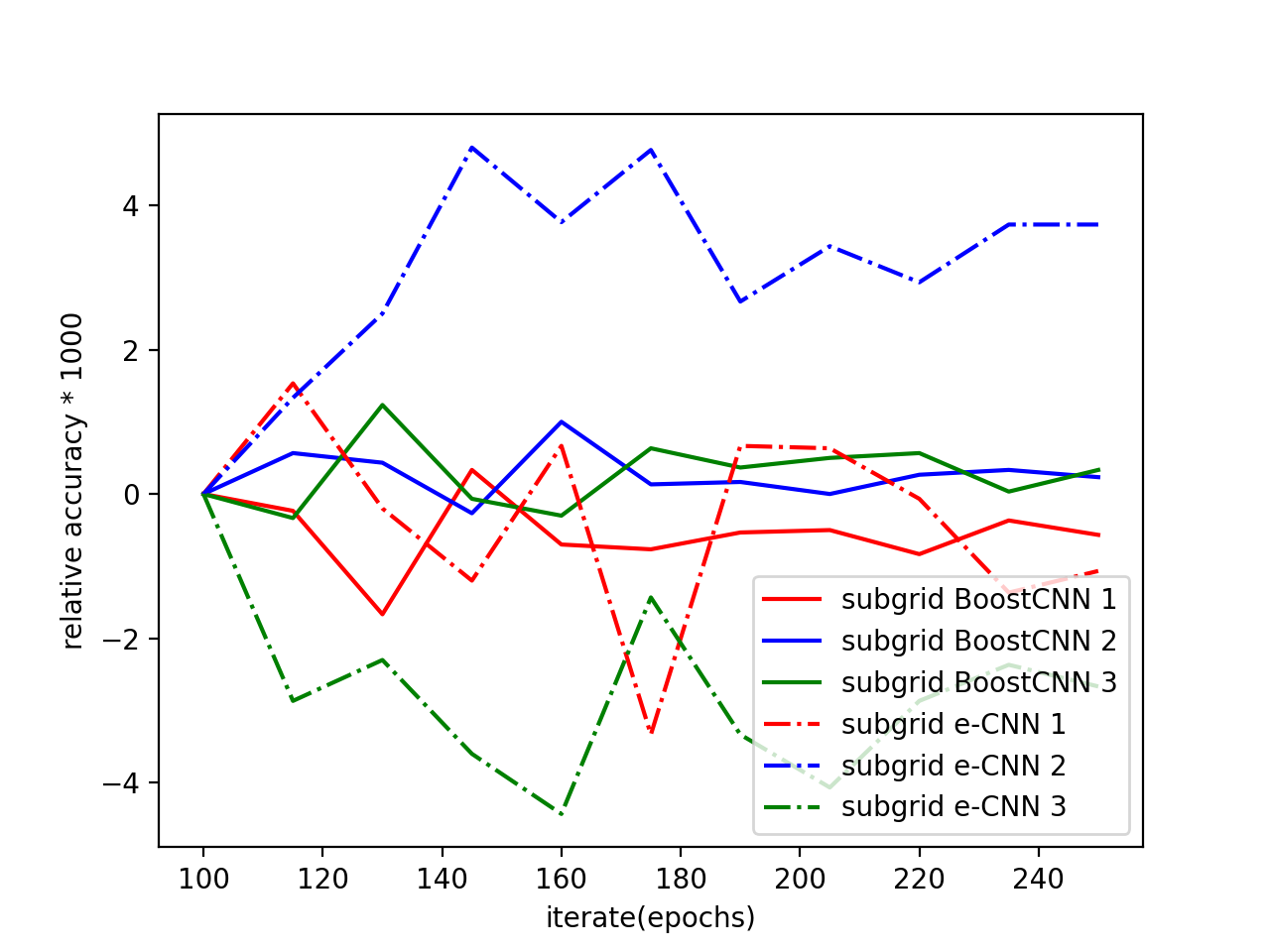}
  \captionof{figure}{Different Seeds}
  \label{fig:18-cifar-seed}
\end{minipage}
\begin{minipage}{.45\textwidth}
  \centering
  \includegraphics[width=\linewidth]{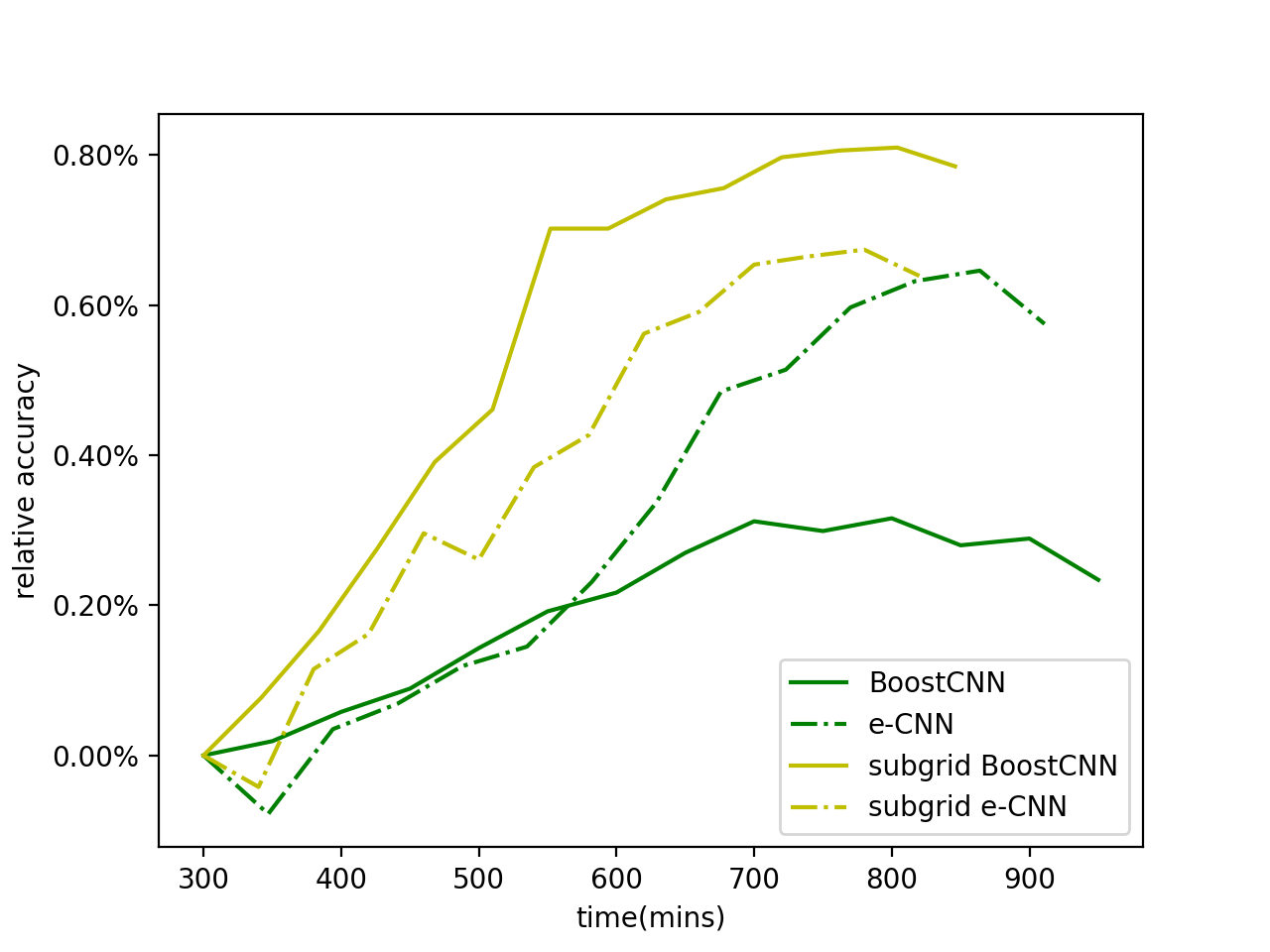}
  \captionof{figure}{ResNet-18 on SVHN}
  \label{fig:18-svhn}
\end{minipage}%
\begin{minipage}{.45\textwidth}
  \centering
  \includegraphics[width=\linewidth]{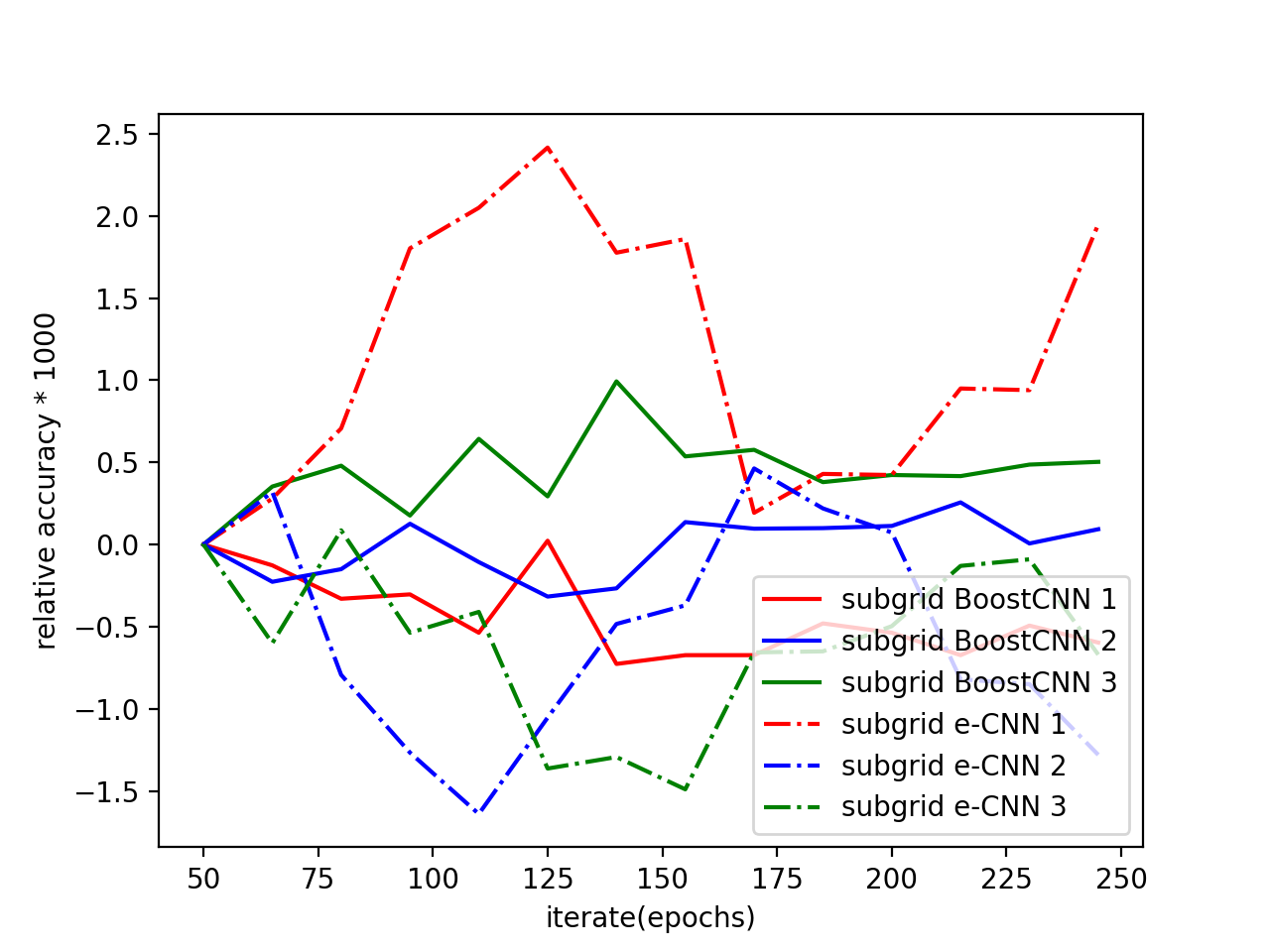}
  \captionof{figure}{Different Seeds}
  \label{fig:18-svhn-seed}
\end{minipage}
\begin{minipage}[t]{.45\textwidth}
  \centering
  \includegraphics[width=\linewidth]{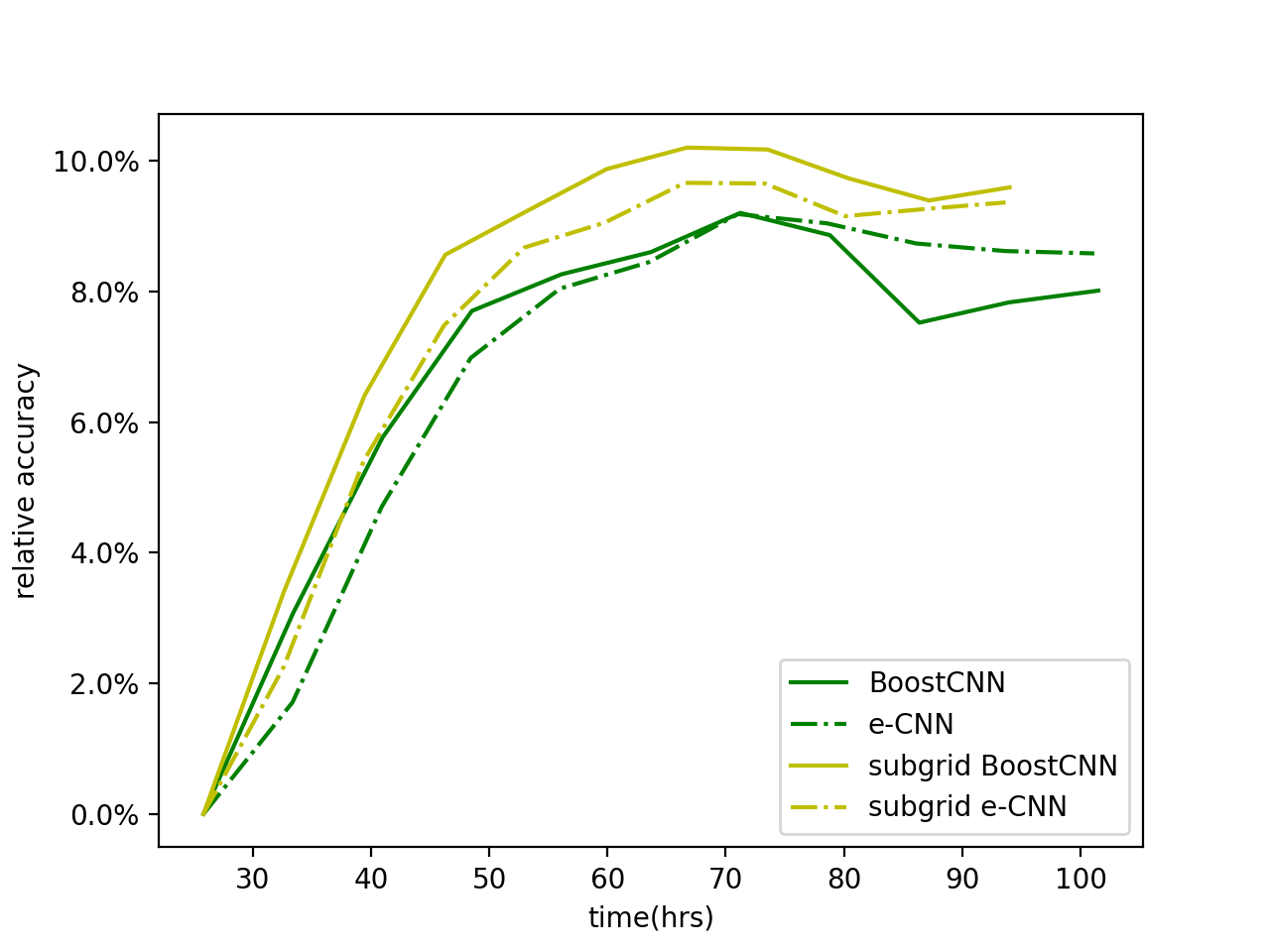}
  \captionof{figure}{ResNet-18 on ImageNetSub}
  \label{fig:18-image}
\end{minipage}%
\begin{minipage}[t]{.45\textwidth}
  \centering
  \includegraphics[width=\linewidth]{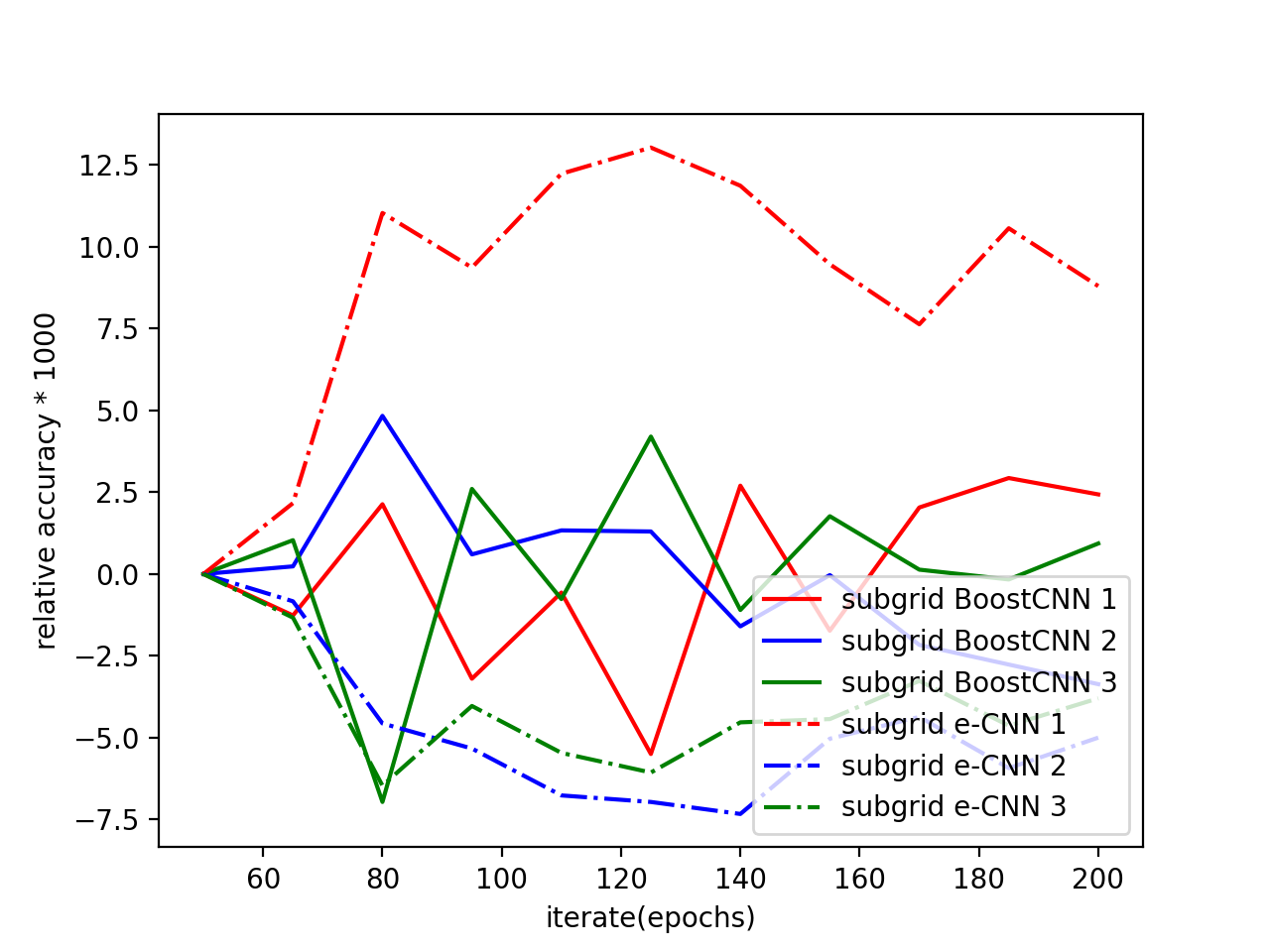}
  \captionof{figure}{Different Seeds}
  \label{fig:18-image-seed}
\end{minipage}
\end{figure*}

We start by applying ResNet-18 as our weak learner for all different ensemble methods. Figures \ref{fig:18-cifar}, \ref{fig:18-svhn} and \ref{fig:18-image} compare the relative performances with respect to single ResNet-18 vs the running time. The solid lines in green and yellow show the relative performances of BoostCNN and subgrid BoostCNN, respectively, while the dotted lines in green and yellow represent the relative performances of e-CNN and subgrid e-CNN, respectively. As shown in these figures, taking the same amount of time, subgrid BoostCNN outperforms all of the remaining algorithms. Furthermore, we observe that subgrid BoostCNN outperforms BoostCNN, and subgrid e-CNN has the same behavior when compared with e-CNN.

In conclusion, the subgrid technique improves the performance of the boosting algorithm. Moreover, Figures \ref{fig:18-cifar-seed}, \ref{fig:18-svhn-seed} and \ref{fig:18-image-seed} depict subgrid BoostCNN and subgrid e-CNN using three different seeds with respect to their averages. The solid and dotted lines in the same color represent the same seed used in corresponding subgrid BoostCNN and subgrid e-CNN. As the figures show, the solid lines are closer to each other than the dotted lines, which indicates that subgrid BoostCNN is more robust with respect to the variation of the seed when compared with subgrid e-CNN. Furthermore, the standard deviations of the accuracy generated by subgrid e-CNN and subgrid BoostCNN are shown in Table \ref{tb:seed}. The standard deviations of the accuracy generated by subgrid e-CNN are significant compared to those of subgrid BoostCNN, which in turn indicates that subgrid BoostCNN is less sensitive to the choice of the seed. Therefore, subgrid BoostCNN is more robust than subgrid e-CNN.

\begin{table}[H]
\centering
\caption{Standard deviation times $10^3$ of the accuracy results by different seeds}
\label{tb:seed}
\resizebox{0.95\columnwidth}{!}{
\begin{tabular}{|l|r|r|}
\hline
         & subgrid BoostCNN & subgrid e-CNN \\ \hline
CIFAR-10 & 0.478         & 2.519         \\ \hline
SVHN     & 0.385         & 0.891         \\ \hline
ImageNetSub & 2.489         & 7.915         \\ \hline
\end{tabular}
}
\end{table}

\begin{figure*}[htbp]
\centering
\begin{minipage}{.45\textwidth}
  \centering
  \includegraphics[width=\linewidth]{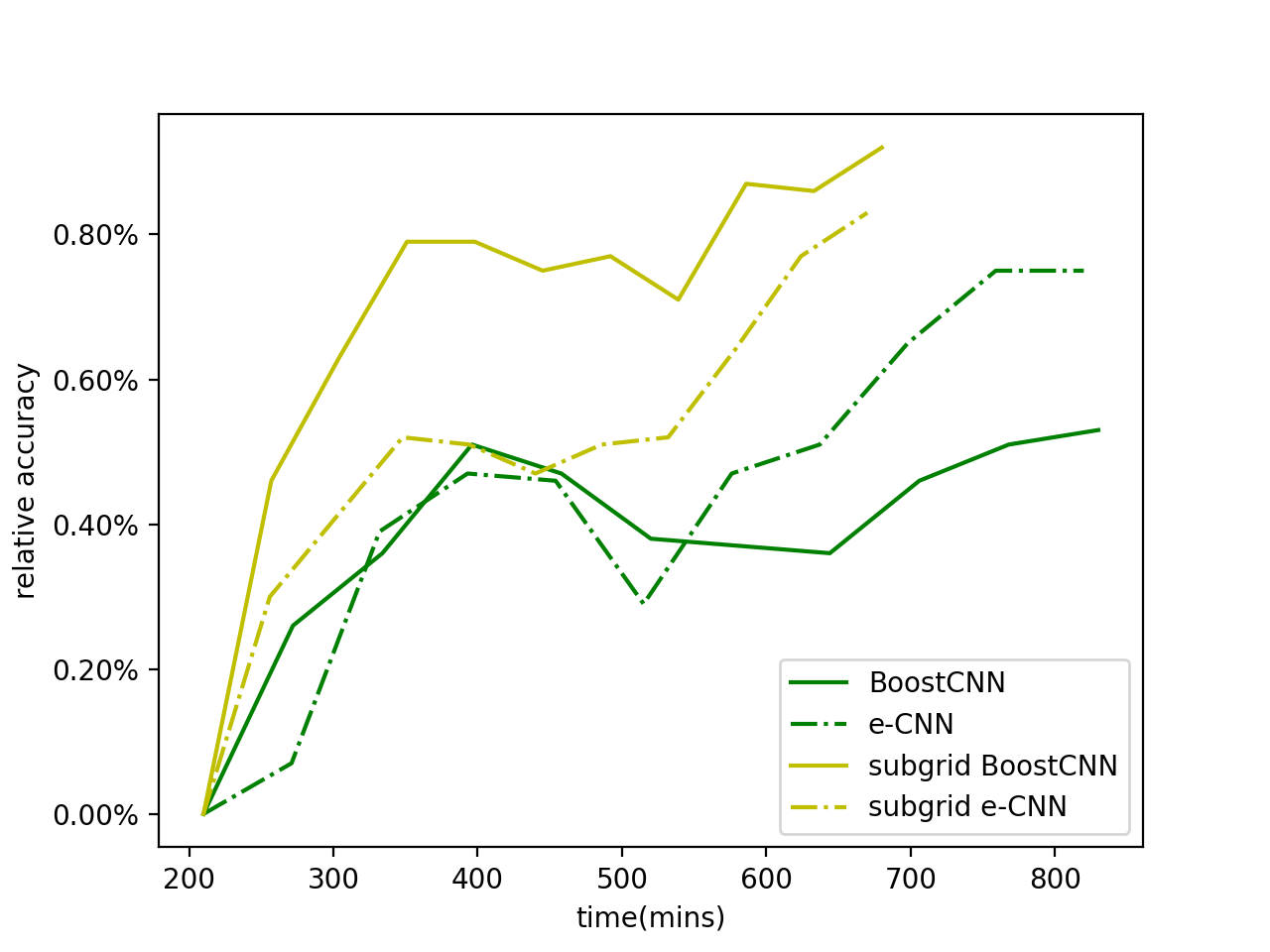}
  \captionof{figure}{ResNet-50 on CIRFAR-10}
  \label{fig:50-cifar}
\end{minipage}%
\begin{minipage}{.45\textwidth}
  \centering
  \includegraphics[width=\linewidth]{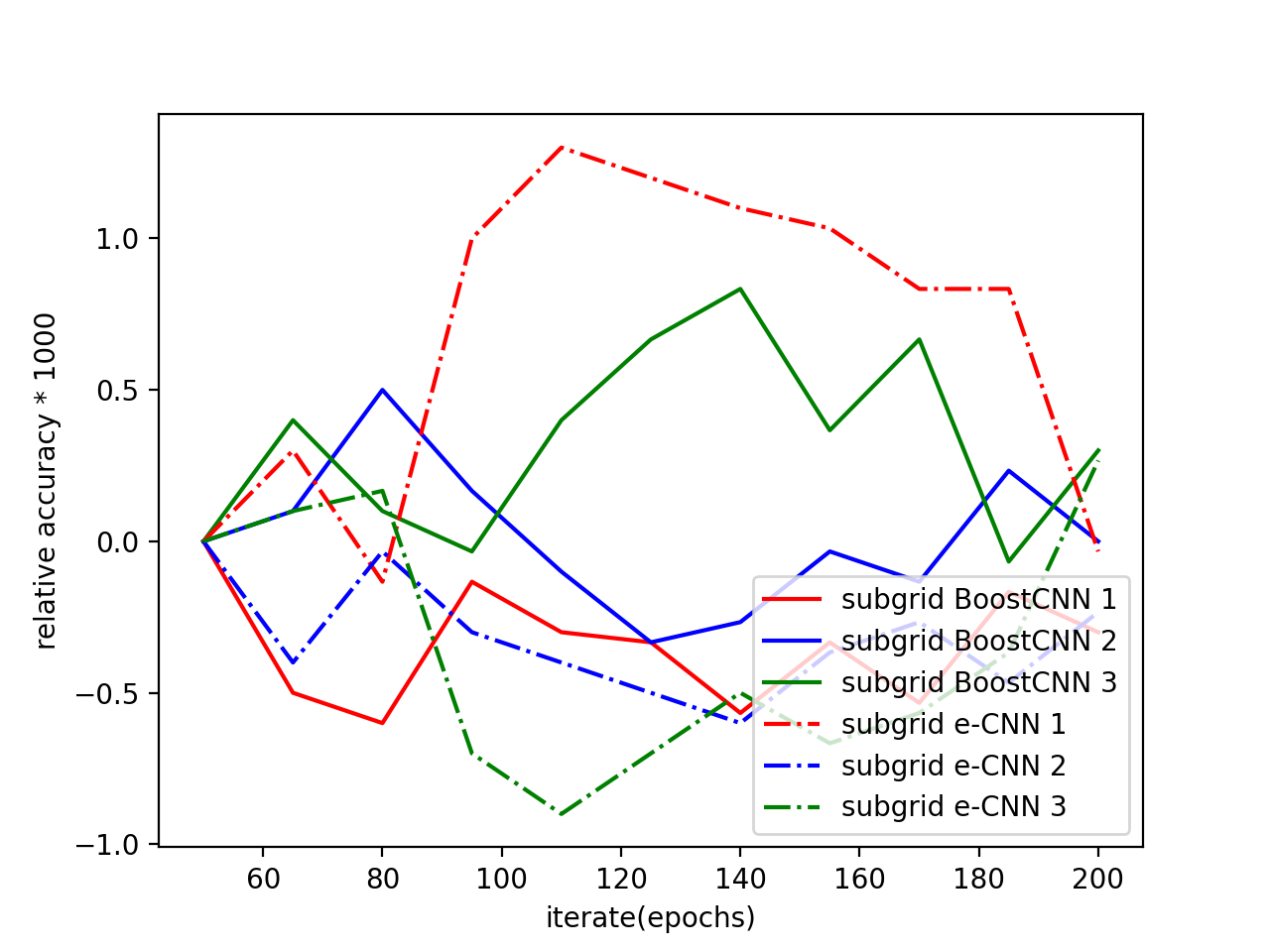}
  \captionof{figure}{Different Seeds}
  \label{fig:50-cifar-seed}
\end{minipage}
\begin{minipage}[t]{.45\textwidth}
  \centering
  \includegraphics[width=\linewidth]{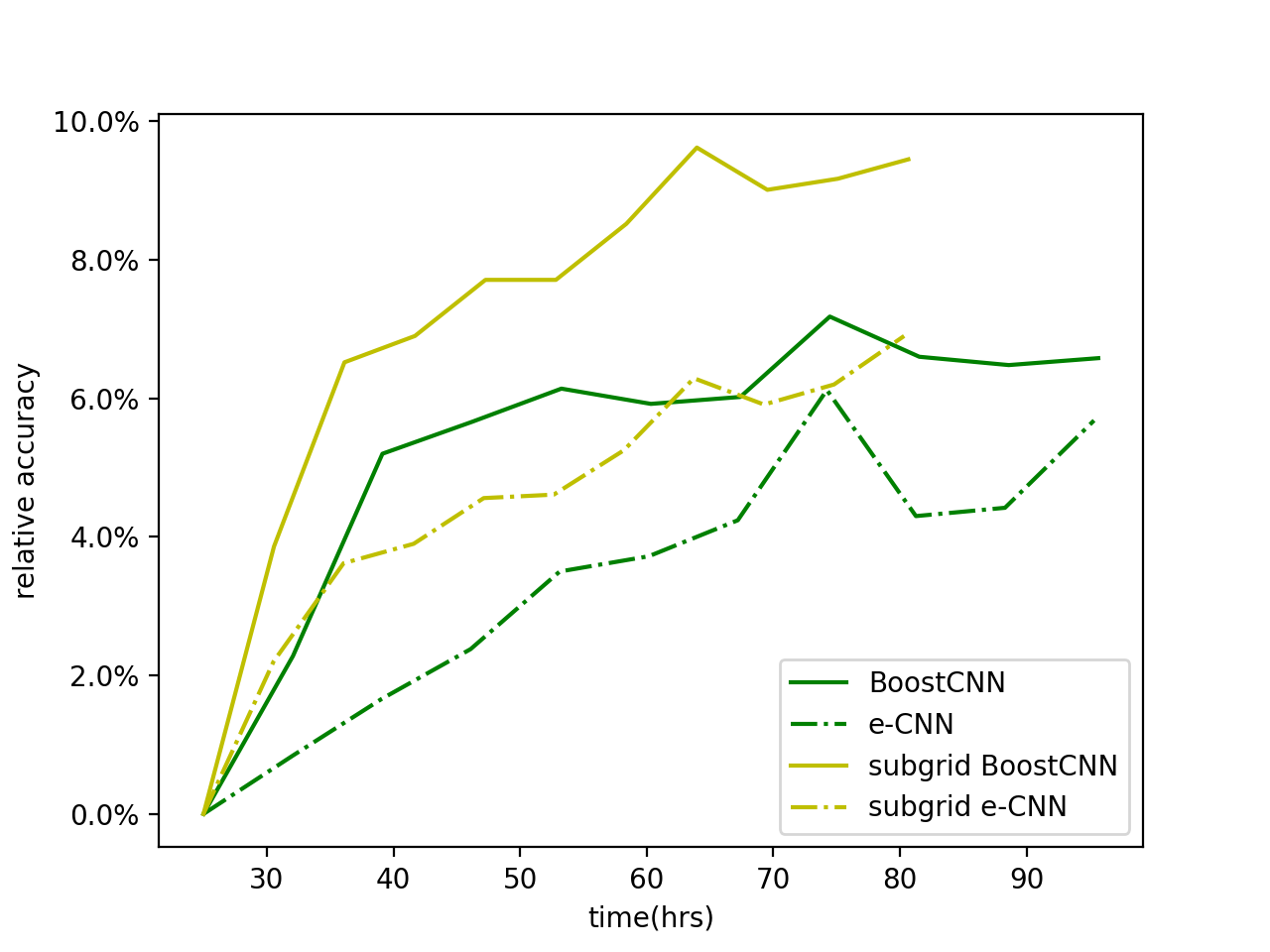}
  \captionof{figure}{ResNet-50 on ImageNetSub}
  \label{fig:50-image}
\end{minipage}%
\begin{minipage}[t]{.45\textwidth}
  \centering
  \includegraphics[width=\linewidth]{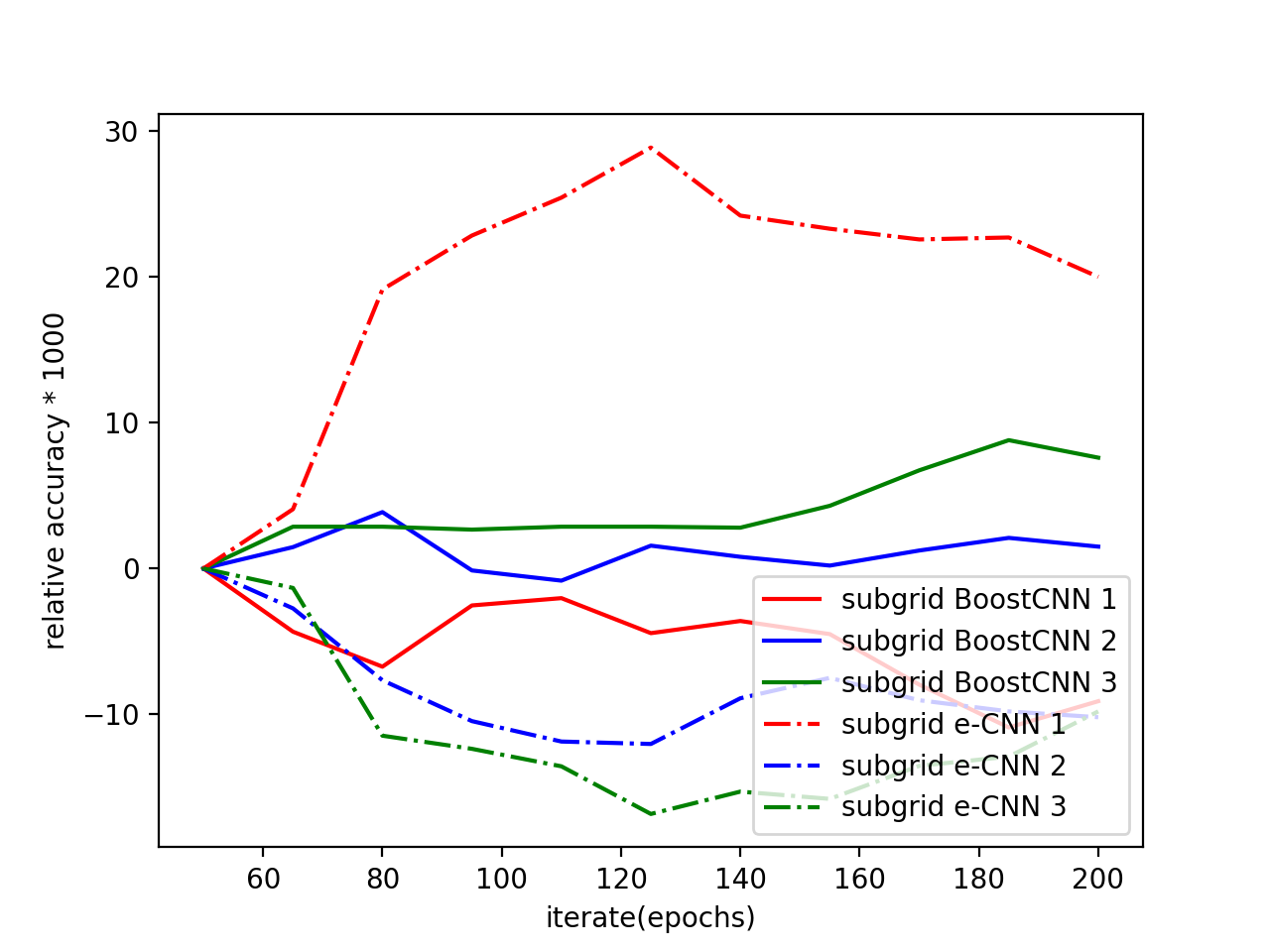}
  \captionof{figure}{Different Seeds}
  \label{fig:50-image-seed}
\end{minipage}
\begin{minipage}{.45\textwidth}
  \centering
  \includegraphics[width=\linewidth]{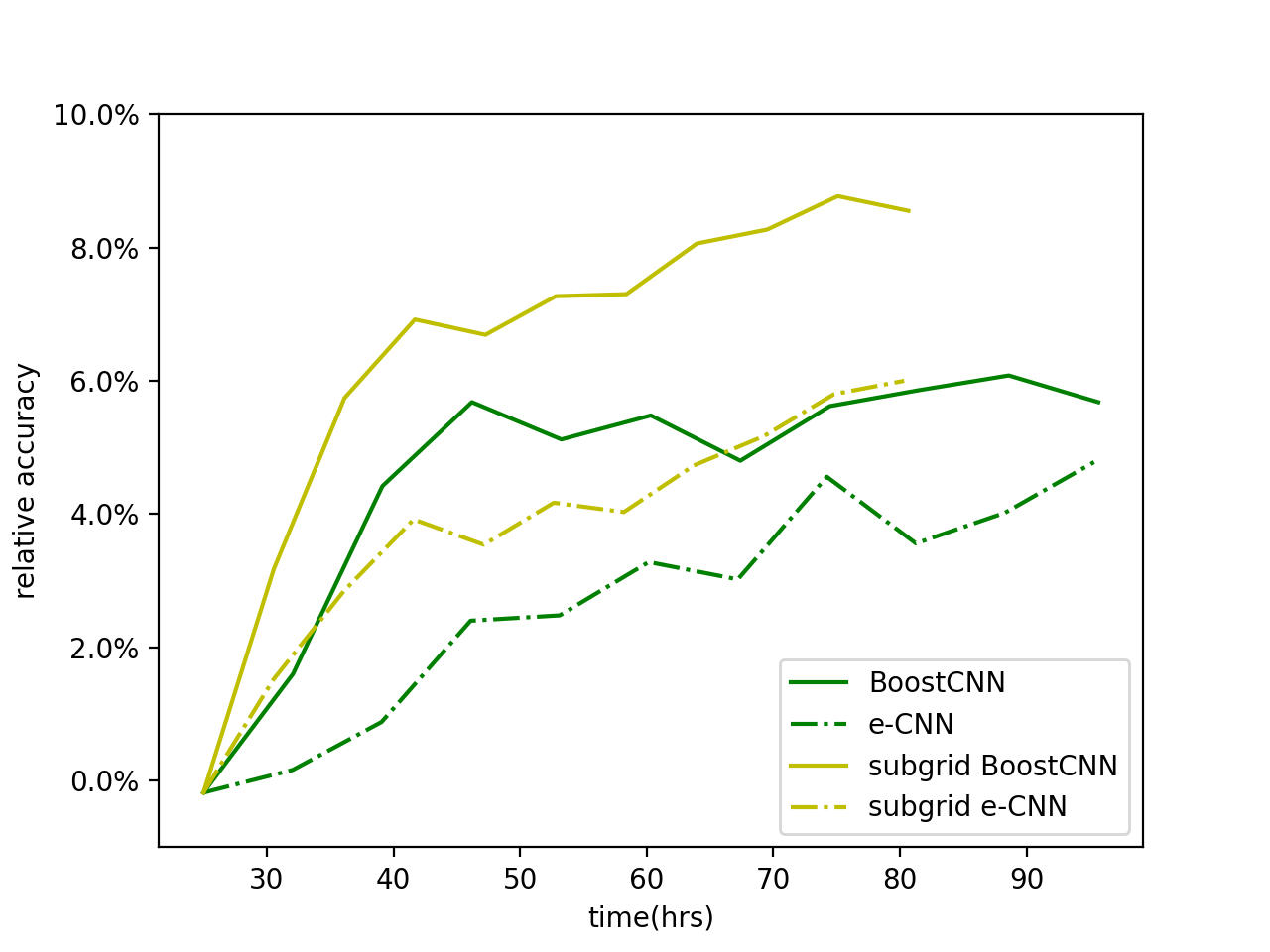}
  \captionof{figure}{ResNet-50 on ImageNetSub compared to ResNet-101}
  \label{fig:50-100 compare}
\end{minipage}%
\begin{minipage}{.45\textwidth}
  \centering
  \includegraphics[width=\linewidth]{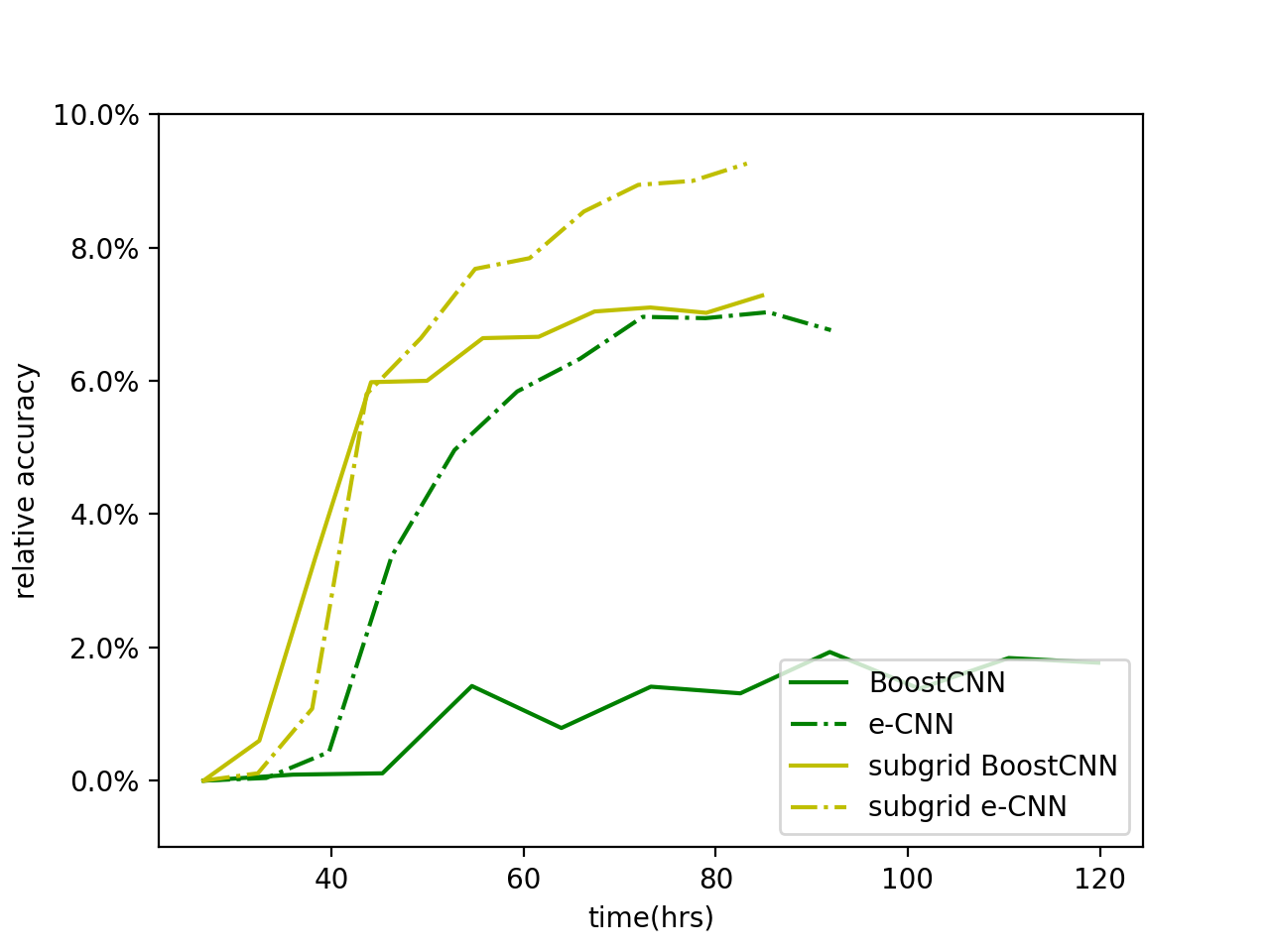}
  \captionof{figure}{ResNet-101 on ImageNetSub}
  \label{fig:101-image}
\end{minipage}
\end{figure*}

Next, we evaluate relative performances of subgrid BoostCNN using ResNet-50 as the weak learner on CIFAR-10 and ImageNetSub datasets with respect to the single ResNet-50. We do not evaluate the relative performances on the SVHN dataset since the accuracy of the single ResNet-50 on the SVHN dataset is over 98$\%$. From Figures \ref{fig:50-cifar} and \ref{fig:50-image}, we also observe the benefits of the subgrid technique. Besides, Figures \ref{fig:50-cifar-seed} and \ref{fig:50-image-seed} confirm that subgrid BoostCNN is more stable than subgrid e-CNN since the solid series are closer to each other compared with the dotted series. 

Furthermore, we establish the relative performances of subgrid BoostCNN using ResNet-50 as the weak learner with respect to the single ResNet-101 in Figure \ref{fig:50-100 compare}. Although single ResNet-101 outperforms single ResNet-50, subgrid BoostCNN using ResNet-50 as the weak learner outperforms single ResNet-101 significantly in Figure \ref{fig:50-100 compare}, which indicates that subgrid BoostCNN with a simpler CNN is able to exhibit a better performance than a single deeper CNN. Lastly, we conduct experiments with ResNet-101 on the ImageNetSub dataset. From Figure \ref{fig:101-image}, we not only discover the superior behaviors of BoostCNN, e-CNN, subgrid BoostCNN and subgrid e-CNN over ResNet-101 as we expect, but also observe the benefit of the subgrid technique.

\section{Conclusion}

Our results show that Subgrid BoostCNN consistently achieves higher accuracy and lower variance than its non-subgrid counterparts. Specifically, with 10 ResNet-18-based weak learners, Subgrid BoostCNN outperforms both BoostCNN and e-CNN across all datasets for the same total training time. It improves accuracy by up to 12.10\% over the base CNN and 4.19\% over BoostCNN, while reducing sensitivity to random initialization. Standard deviation analysis further reveals Subgrid BoostCNN's robustness, especially when compared to subgrid e-CNN, which suffers from higher variance. Additionally, we find that Subgrid BoostCNN generalizes well across architectures. Using ResNet-50 as the base learner, it still outperforms deeper single CNNs like ResNet-101. This suggests that Subgrid BoostCNN can deliver better accuracy even with shallower models, which is particularly beneficial for large-scale or resource-constrained scenarios. In summary, Subgrid BoostCNN effectively balances efficiency, accuracy, and robustness. It offers a scalable and generalizable solution for ensemble learning in vision tasks, making it a compelling alternative to both deep single models and conventional ensemble methods.

% \vspace{5pt}

% \newpage
\clearpage
% \section*{References}
\bibliographystyle{IEEEtran}
\bibliography{references}

\end{document}